\documentclass[letterpaper, 10 pt, conference]{ieeeconf}  
\usepackage{graphicx}
\usepackage{xspace}
\usepackage{tikz}
\usepackage{subfig}

\IEEEoverridecommandlockouts                              

\newcommand\copyrighttext{%
  \footnotesize This work has been submitted to the IEEE for possible publication. Copyright may be transferred without notice, after which this version may no longer be accessible.}
  
\newcommand\copyrightnotice{%
\begin{tikzpicture}[remember picture,overlay]
\node[anchor=south,yshift=10pt] at (current page.south) {\fbox{\parbox{\dimexpr\textwidth-\fboxsep-\fboxrule\relax}{\copyrighttext}}};
\end{tikzpicture}%
}

\title{\LARGE \bf
SwarmUS: An open hardware and software on-board platform for swarm robotics development  
}

\begin{document}

\author{
É. Villemure, P. Arsenault, G. Lessard, T. Constantin, H. Dubé, L.-D. Gaulin, X. Groleau,\\ S. Laperrière, C. Quesnel and F. Ferland
\thanks{All authors are with the Department of Electrical Engineering and Computer Engineering of the University of Sherbrooke (2500 boulevoard de l'Université, Sherbrooke, Québec (Canada), J1K-2R1). F. Ferland is an IEEE member. Contacts for the authors:
\{Etienne.Villemure, Philippe.Arsenault2, Gabriel.Lessard, Thierry.Constantin, Hubert.Dube, Louis-Daniel.Gaulin, Xavier.Groleau, Samuel.Laperriere, Charles.Quesnel, Francois.Ferland\}@usherbrooke.ca.} 
}



\maketitle
\thispagestyle{empty}
\pagestyle{empty}

\newcommand{\UWB}{Ultra-Wideband\xspace}
\newcommand{\wifi}{Wi-Fi\xspace}
\newcommand{\HB}{\textit{Hiveboard}\xspace}
\newcommand{\BB}{\textit{Beeboard}\xspace}
\newcommand{\HBs}{\textit{Hiveboards}\xspace}
\newcommand{\BBs}{\textit{Beeboards}\xspace}
\newcommand{\HC}{\textit{HiveConnect}\xspace}
\newcommand{\HM}{\textit{HiveMind}\xspace}
\newcommand{\HMB}{\textit{HiveMindBridge}\xspace}
\newcommand{\HA}{\textit{HiveAR}\xspace}
\newcommand{\MCU}{microcontroller\xspace}
\newcommand{\MCUs}{microcontrollers\xspace}
\newcommand{\pioneer}{Pioneer 2DX\xspace}
\newcommand{\turtlebot}{TurtleBot3 Burger\xspace}

\newcounter{equationcounter}
\setcounter{equationcounter}{1}
\newcommand{\equationref}{
    \eqno{(\arabic{equationcounter})} \addtocounter{equationcounter}{1}
}


\begin{abstract}
Real life implementations of distributed swarm robotics are rare. 
The standardization of a general purpose swarm robotics platform could greatly accelerate swarm robotics towards real life implementations. 
The SwarmUS platform is an open-source hardware and software on-board embedded system designed to be added onto existing robots while providing them with swarm features, thus proposing a new take on the platform standardization problem.
These features include a distributed relative localization system based on \UWB, a local communication system based on \wifi and a distributed coordination system based on the Buzz programming language between robots connected within a SwarmUS platform. 
Additionally, a human-swarm interaction mobile application and an emulation of the platform in the Robot Operating System (ROS) is presented. 
Finally, an implementation of the system was realized and tested on two types of robots : a \turtlebot and two \pioneer. 
\end{abstract}

\bigbreak


\IEEEpeerreviewmaketitle

\copyrightnotice

\section{INTRODUCTION} 
Swarm robotics is the study of multi-robot systems orchestrated with swarm intelligence to perform certain tasks. 
Even if the entertainment industry has shown centrally controlled swarms of drones in action \cite{waibel_drone_2017}, the main focus of the swarm robotics field revolves around distributed system to remove any single point of failure.
Those distributed systems show promising robustness, fault-tolerance and flexible characteristics compared to single robots when achieving specific tasks \cite{m_dorigo_swarm_2021}. 
However, swarms of robots are rarely seen in real life applications.
According to Nedjah and Junior \cite{nedjah_review_2019}, it is mainly caused by a lack of standard methodologies and the use of generic hardware and software platforms in swarm robotics development.
Establishing those standards could greatly accelerate research towards real life applications.
Multiple robotics platforms are available such as E-Puck \cite{mondada_e-puck_2009}, Kilobot \cite{m_rubenstein_kilobot_2012}, and R-One \cite{j_mclurkin_robot_2014}. However, these platforms were not designed to test real-life scenarios where a task is performed in an environment not suitable for small wheeled robots. 
Additionally, there is a lack of compatibility between the communication system and the coordination system of the majority of those available platforms \cite{nedjah_review_2019}. Therefore, it hinders the development of heterogeneous swarms by the community.

The open source SwarmUS platform proposes a different approach to the hardware standardization problem: giving existing non-swarm dedicated robot models capabilities to work as a distributed swarm. 
These provided swarm capabilities are comprised of a distributed coordination system, a local communication network and a relative localization system. 
This paradigm takes advantage of the broad spectrum of robot models that have all the necessary sensors, computation power, and actuators to perform more realistic swarm applications. It allows the development of distributed heterogeneous swarm while using a common hardware and software platform for the swarm related features. Additionally, researchers can retrofit this platform on the robots they already own. Therefore, it facilitates the inclusion of new researchers in the swarm robotics community while taking advantage of their expertise on their own equipment.

To the best of our knowledge, this idea of adding swarm capabilities to generic robots has been explored only once in the literature \cite{chamanbaz_swarm-enabling_2017}.
The SwarmUS platform differs from that work mainly from the custom hardware platform, the use of \wifi instead of Zigbee for higher bandwidth, the use of Buzz \cite{pinciroli_buzz_2016} as a swarm coordination system, its integration with the Robot Operating System (ROS) \cite{quigley_ros_2009} and the included relative localization feature.
We believe that the addition of these features may offer a viable solution towards widespread adoption and standardization.




This paper is organized as follows: Sect. \ref{system_description} presents the SwarmUS platform and describes each components of the system from a hardware and software perspective. 
Sect. \ref{results} presents results from an experimental validation of the main features of the platform as observed from its integration on three robots (two modified \pioneer  and one modified \turtlebot) executing a "follow the leader" task.
Section \ref{discussion} discusses the current performances and limitations of the original prototype.




\section{SYSTEM DESCRIPTION}\label{system_description}



\begin{figure}[t]
    \centering
    \includegraphics[width=0.5\textwidth]{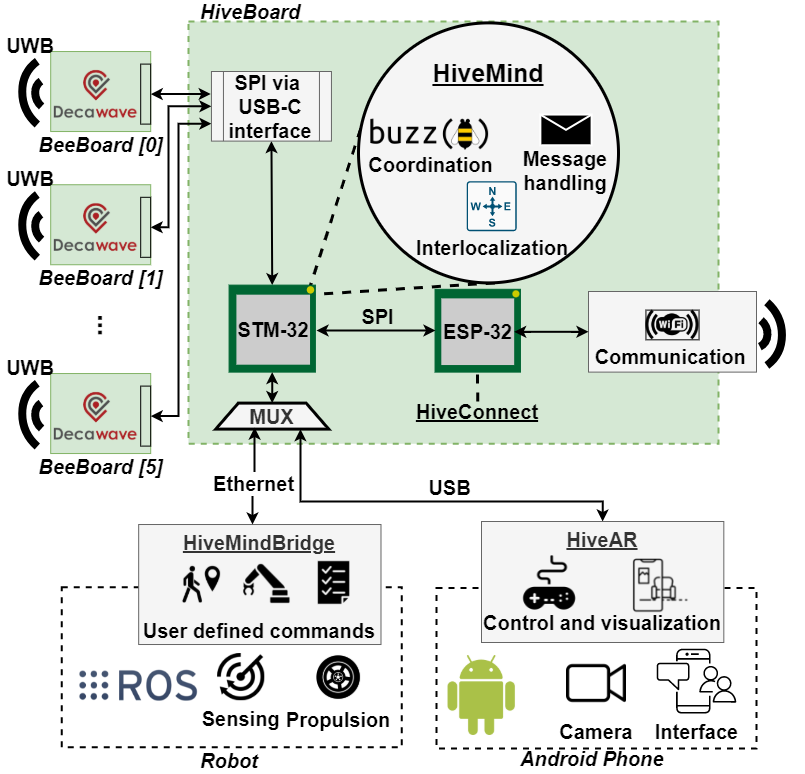}
    \caption{High-level diagram of the SwarmUS platform illustrating the relations between the \HB, the \BBs, the host devices (Example: robot or phone) and their internal systems. 
    }
    \label{fig:global}
\end{figure}

The SwarmUS platform is an open hardware and software on-board system composed of multiple software libraries and two types of electronic circuit boards: the \HB and the \BB. 
Fig. \ref{fig:global} shows the high-level architecture of the SwarmUS platform and how the libraries and circuit boards can be interfaced with a robot or an Android phone for Human-Swarm Interaction (HSI).
Integrating the SwarmUS platform on a robot grants it three general swarm features: coordination, communication, and relative localization.
Both coordination and communication features are implemented on the \HB, a custom made 8-layer printed circuit board (PCB) which is the central hardware component of the platform.

The coordination feature manages the swarm intelligence and the behaviors of an individual robot with all the other robots equipped with the platform. 
,The implementation of the swarm behaviors is handled by the Buzz programming language \cite{pinciroli_buzz_2016} which was specifically designed to program a swarm of robots. 
Each \HB has a Buzz Virtual Machine (BVM) where the Buzz scripts are executed.

The communication feature creates a local network between all the \HBs using \wifi. 
This feature supports the necessary needs in communication for the coordination feature while also enabling robot to robot communication.

The SwarmUS platform can create a standalone swarm of \HBs. 
However, since the \HB has two wired interfaces (USB and Ethernet) to communicate with external devices, a swarm can be formed from any such devices.
Theses devices are assigned as hosts, and can be heterogenous robots or Android phones or tablets. 
An Android device can thus be connected through USB to a \HB, join the swarm and then be used to visualize and send commands to the swarm, enabling HSI.
In the scope of this project an Android application, named \HA, has been developed to configure and monitor the state of the swarm.

Because of the level of abstraction between the \HB and the host device, the swarm development can be disconnected from the development of robot features like sensing and navigation. 
Therefore, the focus of the swarm robotics community could be put around standardizing a swarm platform without worrying about the different mechanical base and computational capabilities of robots.

The last feature, the relative localization system, allows robots equipped with the system to measure their relative distance and bearing from each other. 
This type of relative measurement is mandatory for multiple swarm behaviors~\cite{bishop_target_2007} ~\cite{shiell_bearing-only_2016}.  
These measurements are obtained through the \UWB (UWB)
technology of the \BBs. UWB offers more range and has more environmental flexibility (ex: not dependent on ambient lighting) compared to other local measurement systems used in swarm robotics \cite{a_kohlbacher_low_2018}. 
Additionally, the system is independent from a centralized system like GPS
or overhead cameras for smaller robots
to work, giving more robustness to the swarm.
As show in Fig.\ref{fig:global}, at least one \BB needs to be connected to a \HB via a USB-C interface in order to provide those measurements. 
A single \BB connected to a \HB  gives the relative distance between every other \BBs in line of sight.
If there is at least three \BBs, it can also give the bearing angle. 
Up to six \BBs can be connected to a single \HB to enhance the measurements. 

The \HB contains an STM32 \MCU unit (MCU) which is responsible of the coordination feature, thus executing the BVM and the Buzz scripts. 
It also has a message management system to handle the messages from the other \HBs coming from the \wifi network and from the connected host device.
Lastly, it also contains the relative localization algorithm to calculate the range and the bearing of other robots based on the \BB data. 
All the software components, including the low level hardware interfaces, are combined under the same firmware called the \HM.
The \HB also has an ESP32 module interfacing with the STM32 to manage \wifi networking.
The ESP32 module has is own firmware called the \HC.

To enable communication with the \HB, a C++ library for the host devices, the \HMB, has been developed.
In the scope of the project, the \HMB has only been implemented in a ROS package for devices with an Ethernet interface, but it could be implemented on any C++ platform.  

To ease development with the SwarmUS platform, the same code that runs on a \HB can be cross compiled as a ROS node on any Linux-based system. 
The low level hardware interfaces, the \wifi network and the relative localization system can be all emulated using ROS nodes, ROS topics, and the Gazebo simulator \cite{n_koenig_design_2004} to provide a fully emulated system.
Therefore, it facilitates the transition from simulation development to real world integration, which is an important feature for a standardized platform \cite{m_dorigo_swarm_2021}. 

In the end, once the SwarmUS platform is connected to a host device, the developer only needs to write the Buzz scripts in the \HM.
There, the developer needs to define callback functions called between the \HB and its host device in the \HM and the \HMB.
Finally, they only need to ensure that the host device reacts desirably to these callbacks
by performing the desired actions.

All the hardware and software components of SwarmUS are open source under the MIT License \cite{noauthor_mit_nodate} and can be found on GitHub\footnote{https://github.com/SwarmUS}, along with full documentation\footnote{https://swarmus.github.io/SwarmUS-doc/}.

\subsection{Localization}\label{sys_interloc}
The relative localization feature of the SwarmUS platform is provided by the active \BBs antennas. 
Each \BB harbors a Decawave DW-1000 integrated circuit with a Chiolas ANT110 UWB antenna and can be connected to the \HB via an USB-C connector supporting an SPI bus. 
A system with a \HB-\BB combination can localize another similar system using the two-way ranging (TWR) technique \cite{decawave_aps013_2015}.

\begin{figure}[t]
    \centering
    \includegraphics[width=0.48\textwidth]{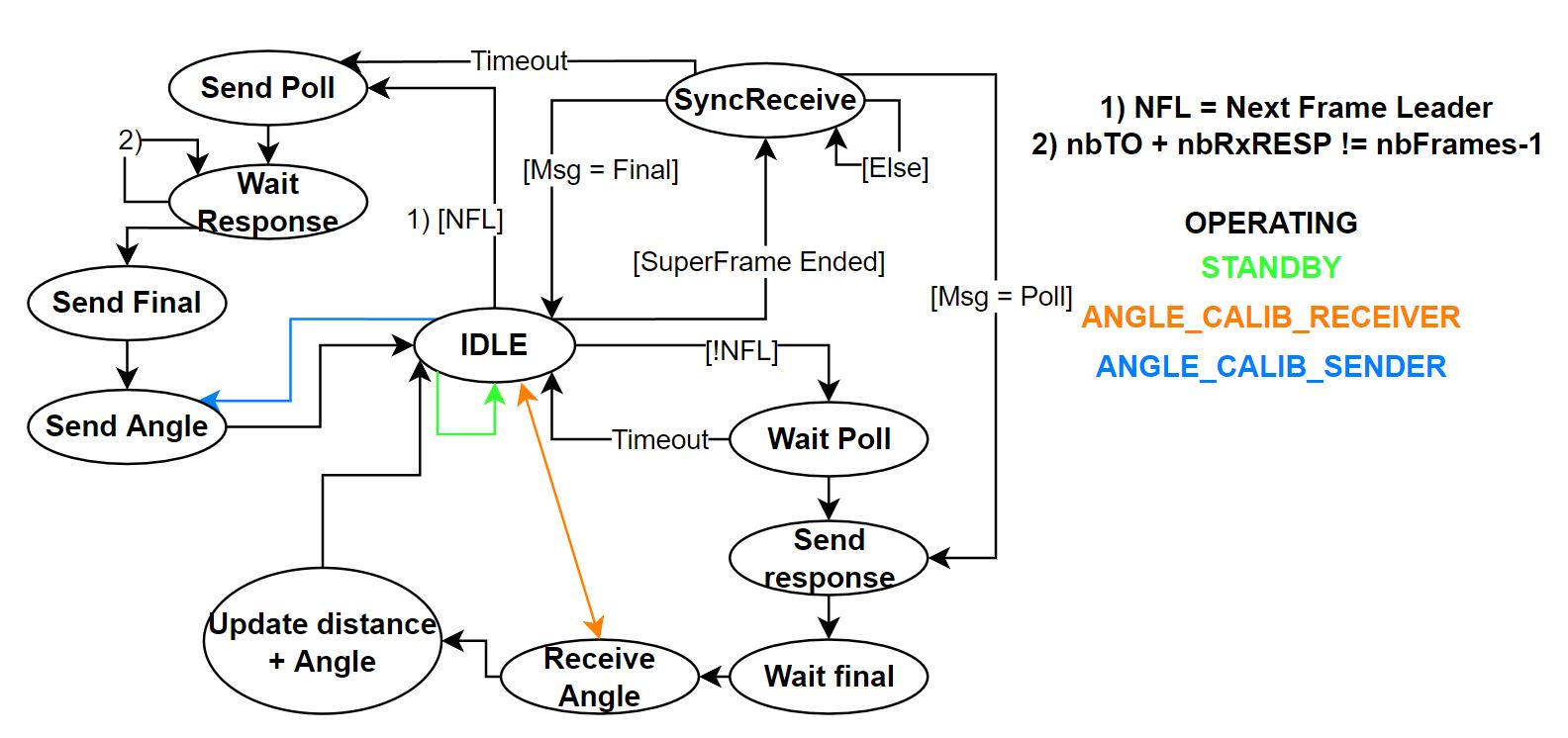}
    \caption{Finite state machine of the localization system}
    \label{fig:FSM}
\end{figure}

TWR consists of exchanging three messages between an initiator and a responder. As the messages travel from one to another, the time of arrival is recorded and passed on with the messages.
Since multiple robots are using TWR in the same air space, message collision might happen.
To prevent this, the SwarmUS platform has a scheduling mechanism where each system has a specific time slot to send its TWR message.
The management of all theses messages and the synchronisation of the SwarmUS platform is done via a finite state machine (Fig. \ref{fig:FSM}). The synchronisation is performed for every robot that joins the swarm and happens during the "SyncReceive" step in FSM the diagram.
However, the number of time slots in the schedule is set during compile time so the maximum scale of the swarm must be known beforehand.

\begin{figure}
    \centering
    \includegraphics[width=0.4\textwidth]{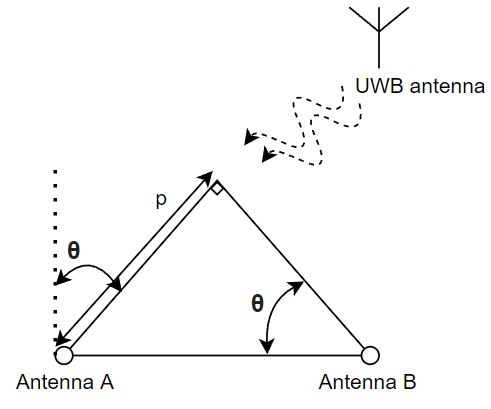}
    \caption{Geometric relation between the difference of phase (p) of two antennas and the bearing angle ($\theta$)
    }
    \label{fig:AOA}
\end{figure}

To determine the bearing measurement, a pair of \BB antennas are necessary to measure the phase difference of arrival.
When a signal is received by two synchronized \BBs, a quadrature system as described in \cite{dotlic_angle_2017} is used to extract a phase difference.
From this difference, an angle can be extracted in a [-90, 90]$^{\circ}$ range as it can be seen in Fig. \ref{fig:AOA}. 
Since a single pair cannot tell the difference between a front and back arrival, a third antenna must be introduced to create an equilateral triangle formation to break the mirror effect, therefore creating three pairs of antennas.
Furthermore, a single antenna pair has difficulties when the absolute value of the angle is high because the calculation of the angle is based on the \textit{arctan}\footnote{https://swarmus.github.io/SwarmUS-doc/sections/reference/Interloc/
how\_it\_works/angle/} function. 

It is imperative to characterize the angular response with a test bench to compensate any misplacement in the position of the antennas and to verify if there are unreachable or inaccurate zones of reception.
Doing so can reveal blind spots of each antenna pair that can be covered by better placed pair.
In the SwarmUS platform, each antenna pair has a weight corresponding to the certainty of the measurement.
The certainty is based on the value angle returned and a line of sight quality indicator (as provided provided by an internal registry of the DW-1000), any measure outside [-30,30]$^{\circ}$ or [150,210]$^{\circ}$ is considered more uncertain.
Proceeding this way, the false values caused by measurement errors are minimized and the best values are maximized before the angle is returned.

\subsection{Coordination} \label{subsec:Coordination}

The SwarmUS plateform leverages the Buzz \cite{pinciroli_buzz_2016} language 
to implement swarm behaviors via a virtual machine (BVM).
Since this programming language was tailor-made for controlling robotic swarms, it contains key data structures and functionality to manage groups of robots.
For instance, it can easily be used to manage a shared data table updated by agents in the swarm called the \textit{stigmergy}.
It also provides a \textit{neighbor} type used to store positional information on the robots in the swarm as well as communicate with \textit{broadcast} and \textit{listen} methods. 

\begin{figure}[ht]
    \centering
    \includegraphics[width=0.48\textwidth]{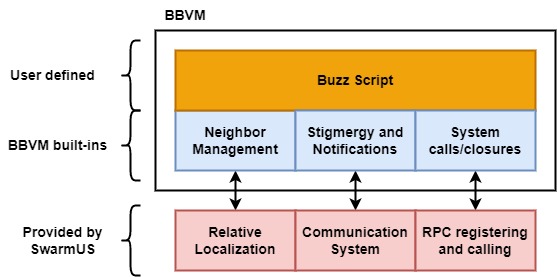}
    \caption{The components constituting the BBVM. The Buzz script is defined by the user, the parts in blue are core systems in the BBVM and the parts in red are the system powering those functionalities on the SwarmUS platform.}
    \label{fig:bbvm_architecture}
\end{figure}

Buzz runs in the BVM, which has a base version to run on Unix systems and also a more lightweight version, the Bitty Buzz Virtual Machine\footnote{https://github.com/buzz-lang/BittyBuzz} (BBVM), targeted for \MCU like ARM Cortexes.
The \HM firmware is built around the BBVM, both for the firmware and its emulation to limit discrepancies between the two implementations.
The integration of the BBVM in the \HM firmware relies on providing implementations to core systems of the BBVM, like relative localization  and communication systems.
Figure \ref{fig:bbvm_architecture} presents the links between the functionalities available in Buzz and their link to the components of the system. 

The firmware also links together the Buzz script and the \textit{host} robot. 
The robot exposes a manifest of actions, like movement or interactions with the environment, that it can perform. This manifest takes the form of a list of callbacks, transmitted between the host and the \HB.
The buzz script controls the robot and this abstraction allows for easy deployment on heterogeneous robots that have different capabilities. 
This manifest of actions can also be used to help the swarm task allocation process based on robot capabilities~\cite{i_budinska_task_2016}.

\subsection{Communication} \label{subsec:system-communication}

The communication system for the platform revolves around a \wifi network, provided by one of the \HBs or by an external access point.
Every board is configured to join the same configured network
to then exchange data.

The communication scheme supports broadcast and unicast communication. 
The broadcast is mainly used to update the \textit{stigmergy} from Buzz whereas the unicast is used to send commands. These commands are sent by other agents and can either target a given agent's \HB or host. 

\begin{figure}[ht]
    \centering
    \includegraphics[width=0.5\textwidth]{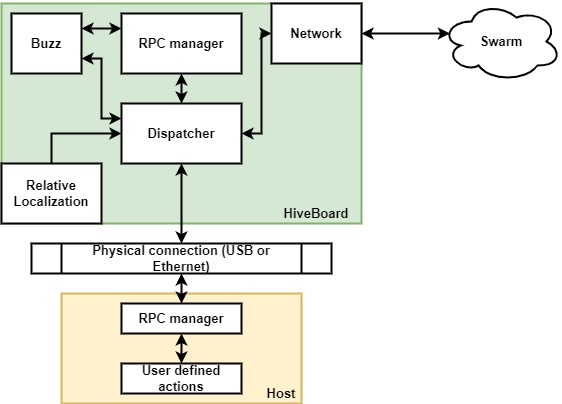}
    \caption{Communication datapath between a host, a \HB and the swarm. Both host and \HB have a dispatcher to forward calls to the proper target. The network box represents the ESP32 and the \HC firmware, which also has an internal dispatcher to translate between agent id and IP addresses for unicast calls.
    }
    \label{fig:communication_rpc}
\end{figure}

The two \MCUs on the \HB will then dispatch the messages to the proper target with a messaging system (Fig \ref{fig:communication_rpc}) built using Google's Protobuf \footnote{https://developers.google.com/protocol-buffers}. 
The system is built around Remote Procedure Calls (RPCs) to either invoke an action or update information, either coming from the rest of the swarm, the host or other systems like the relative localization feature.
 
\subsection{Human-Swarm Interaction (HSI)} \label{subsec:hsi}

\begin{figure}[b]
    \centering
    \subfloat[Agent listing]{
        \includegraphics[width=0.14\textwidth]{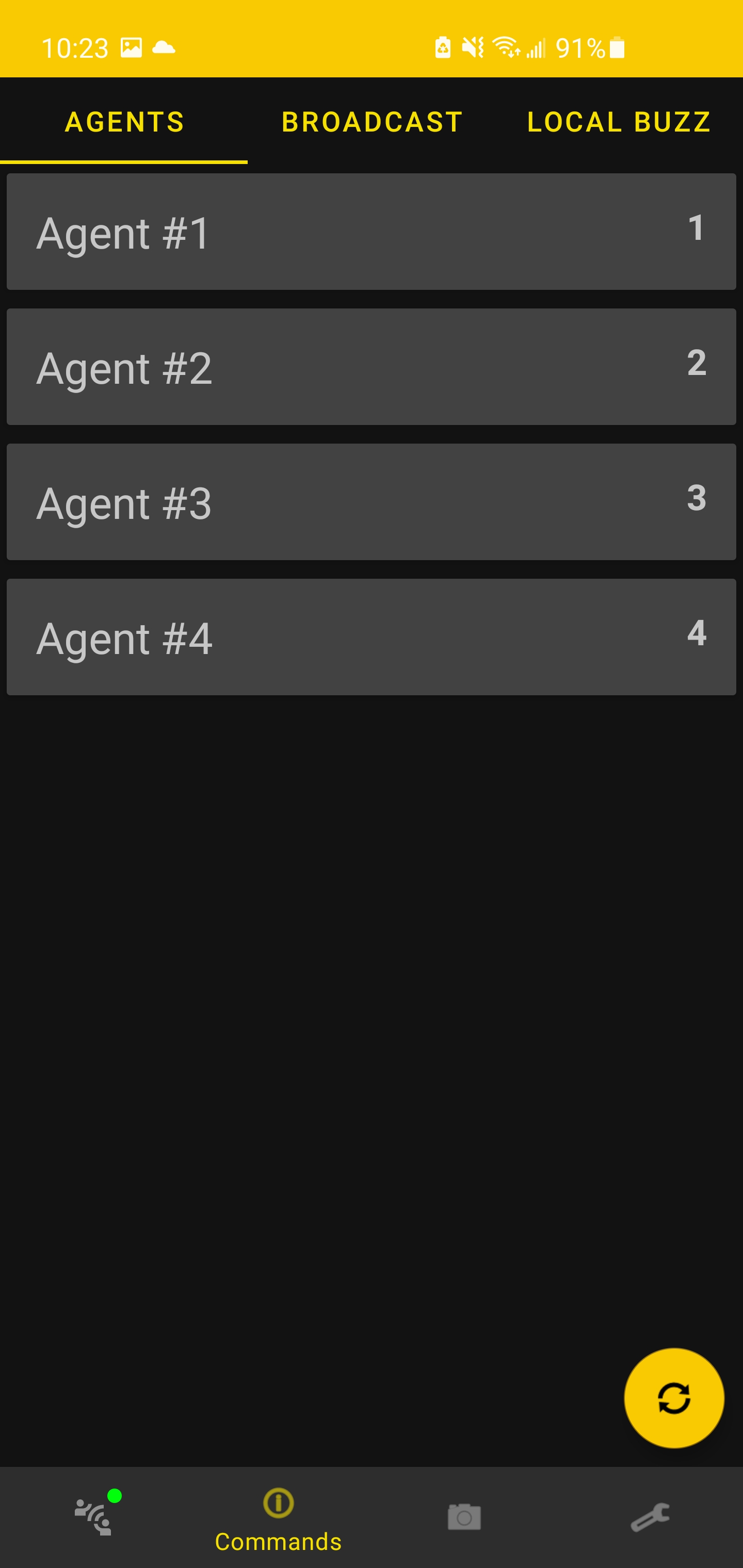}
        \label{fig:interface_controle}
    }
    \subfloat[Available actions to call]{
        \includegraphics[width=0.14\textwidth]{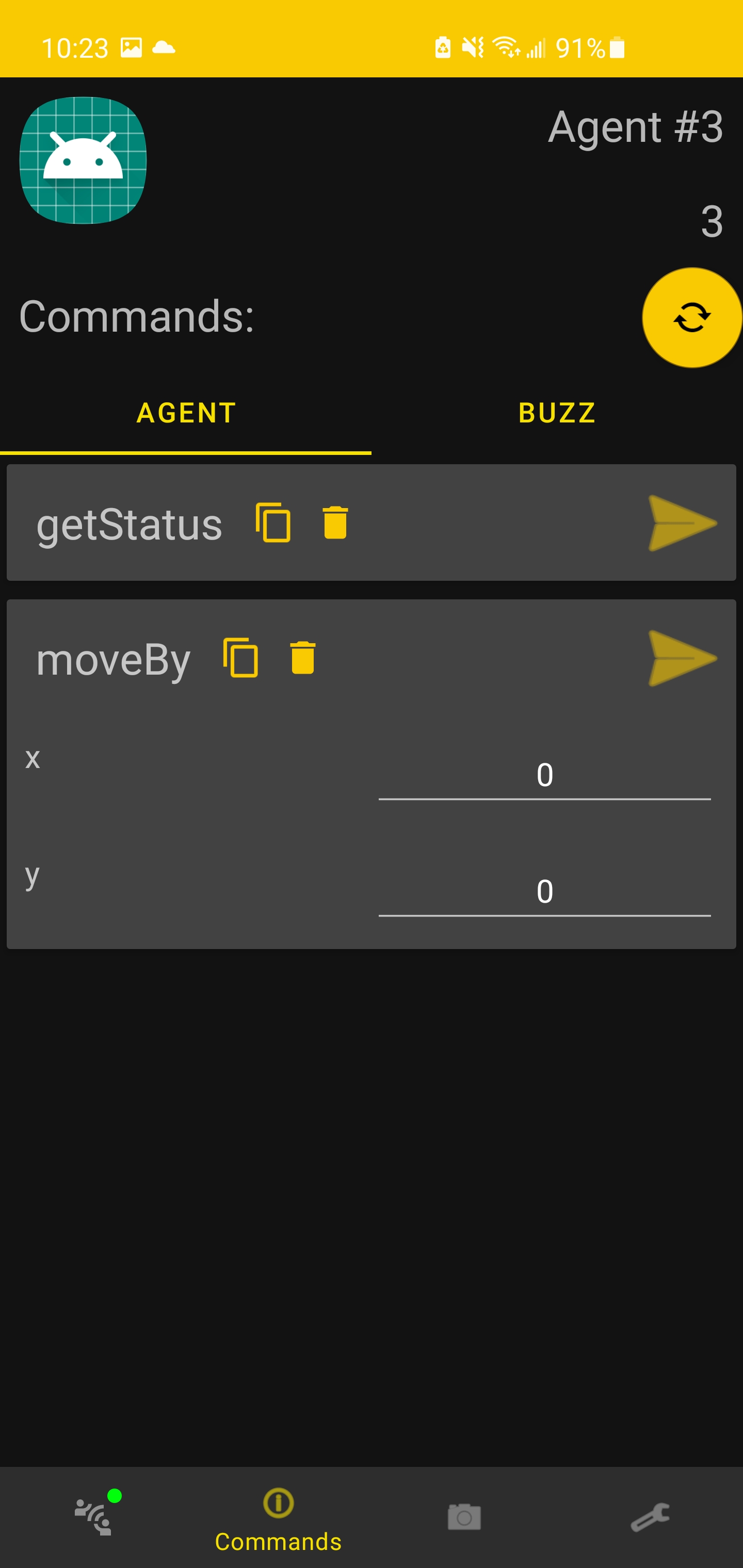}
        \label{fig:interface_agent}
    }
    \subfloat[AR view of a robot represented by an April Tag 6]{
        \includegraphics[width=0.14\textwidth]{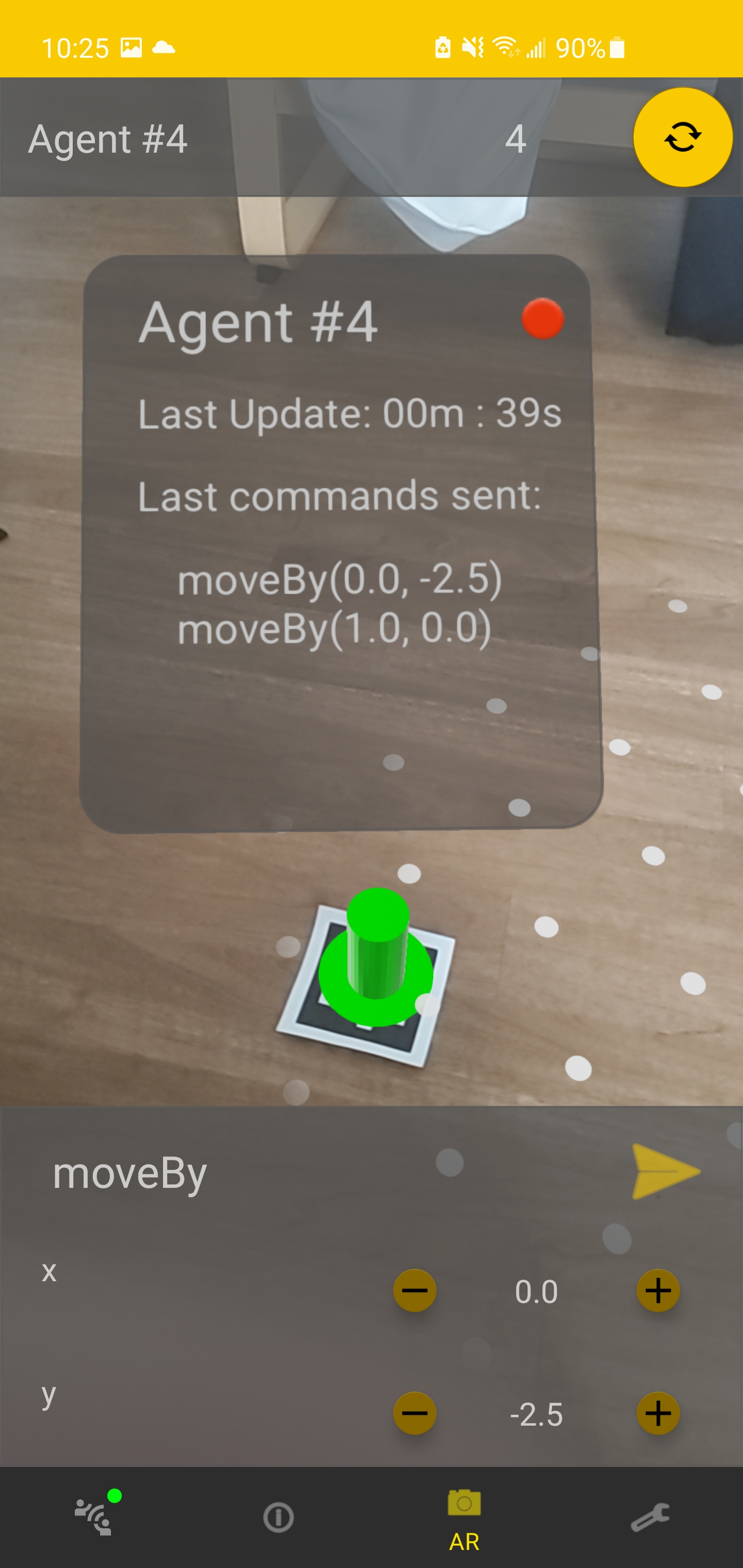}
        \label{fig:interface_agent_ar}
    }
    \caption{The \HA interface}
    \label{fig:HSI}
\end{figure}

The HSI interface, called \HA, revolves around controlling multiple agents, either as a group or individually. 
The interface was built as an Android application that connects to a \HB, making it a member of the swarm. 
Like shown in Fig. \ref{fig:interface_controle} and \ref{fig:interface_agent}, it can view the available actions of an agent, call those actions with parameters and broadcast commands to the swarm. 
It is also responsible to configure some parameters for the \HB like its ID and network configuration. 
It leverages Augmented Reality (AR) features to easily select an agent to view its command, to call them, and to monitor the status of the agent as well as track its movement.
To localize the robot, the AR feature uses the cellphone's camera to find April Tags\footnote{https://github.com/AprilRobotics/apriltag} placed on the robots as it can be observed in Fig. \ref{fig:interface_agent_ar}.

\section{EXPERIMENTAL VALIDATION} \label{results}

\begin{figure}[t]
    \centering
    \subfloat[\HB top view]{
        \includegraphics[width=0.22\textwidth]{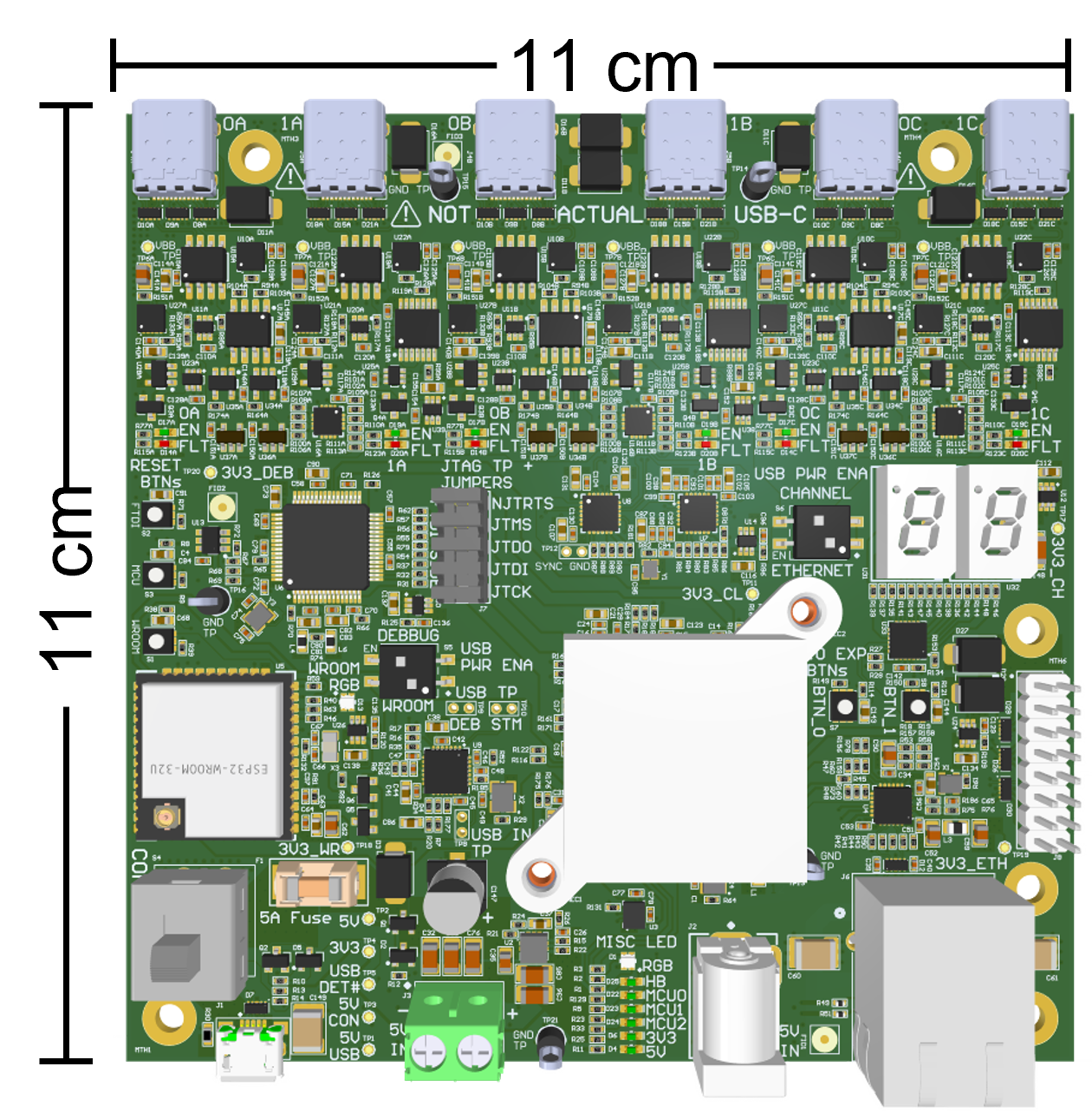}
        \label{fig:HB}
    }
    \subfloat[\BB Top view]{
        \includegraphics[width=0.22\textwidth]{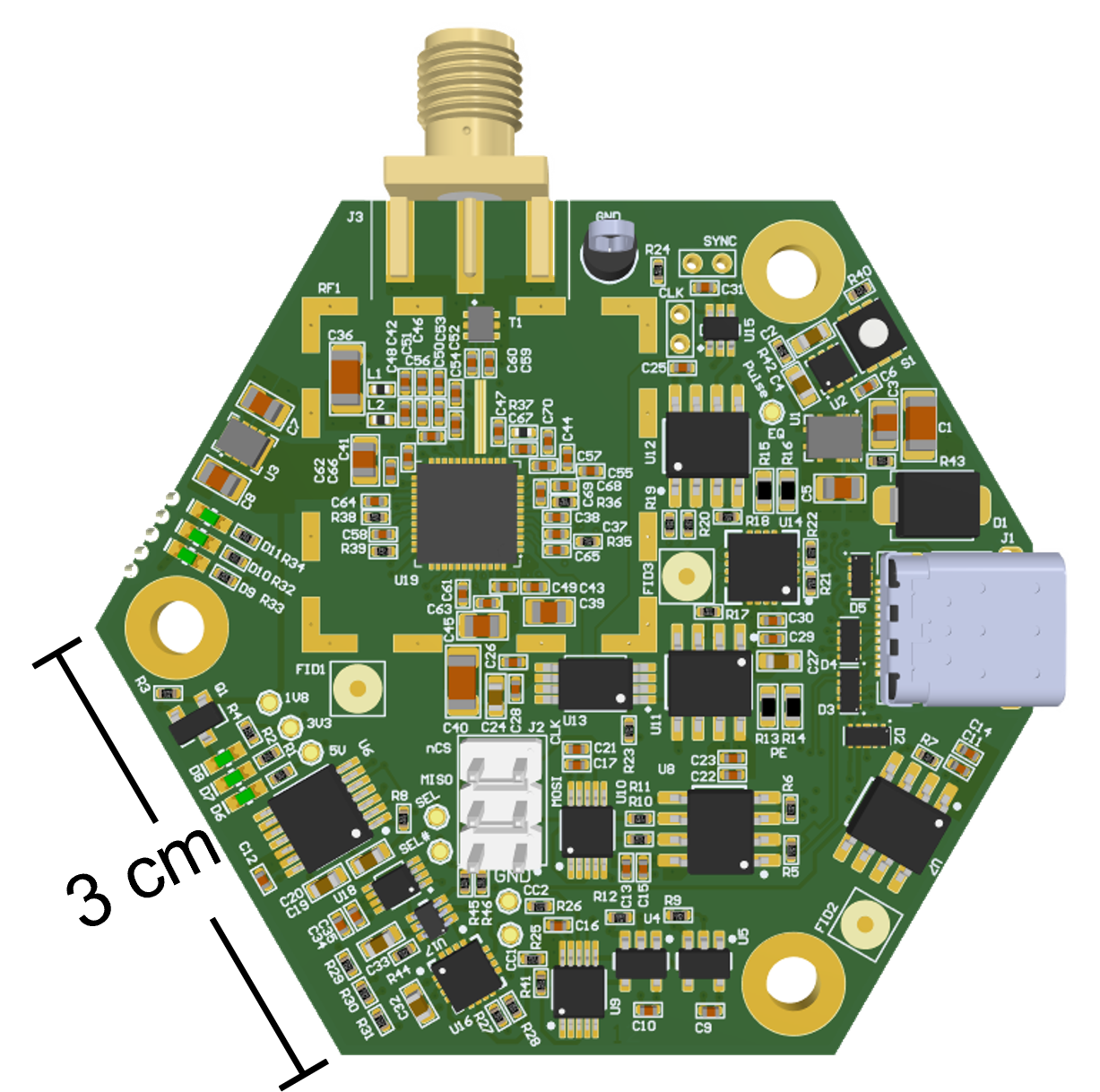}
        \label{fig:BB}
    }
    \caption{SwarmUS boards}
    \label{fig:boards}
\end{figure}

Six \HBs and eighteen \BBs were produced for the experimental validation. The size and top layer of these 8-layer PCB are shown in Fig. \ref{fig:boards}.


The integration of the SwarmUS system was done on two types of robots : a \turtlebot and two \pioneer.
The hardware and software of the differential drive \pioneer have been updated in the scope of this project.
The two \pioneer are equipped with a A2M8 RPLidar lidar, a D400 series Realsense Camera, a 2x30A RoboClaw as their motor controller and a Raspberry Pi 4 4GB as their main computer.
The \turtlebot is a differential drive robot with a small footprint of 13.8 by 17.8 cm equipped with a LDS-01 lidar and has been updated to have a Raspberry Pi 4 4GB.
On the software level, these robots run on a ROS architecturefor safe navigation towards goal positions received by the \HB. Except for the nodes interfacing with sensors and motors, the main implemented nodes are \HMB and \textit{movebase}\footnote{http://wiki.ros.org/move\_base}, responsible for path planning and obstacle avoidance.

\begin{figure}[h]
    \centering
    \includegraphics[width=0.48\textwidth]{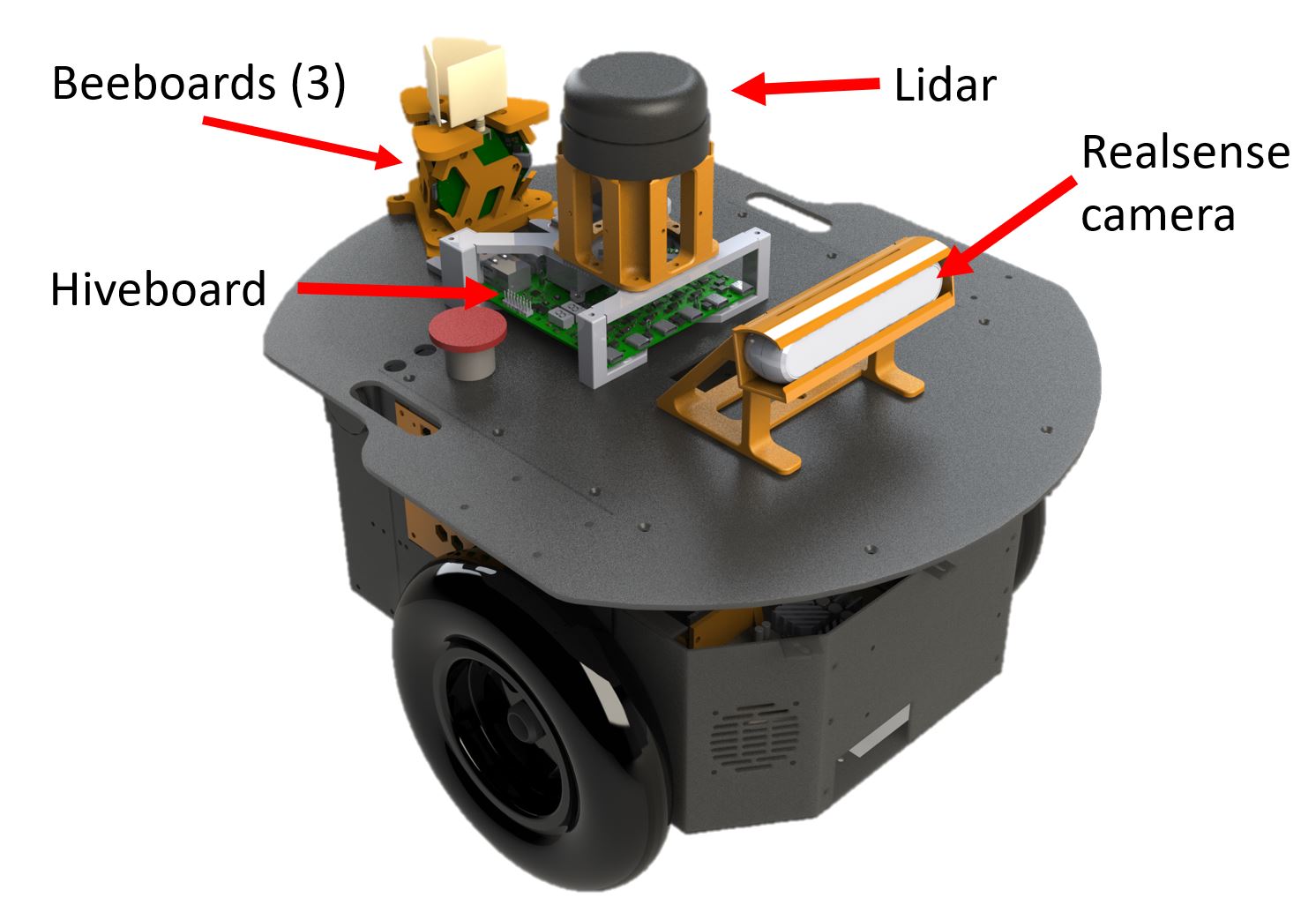}
    \caption{A \pioneer with a SwarmUS system} 
    \label{fig:robot}
\end{figure}

Figure \ref{fig:robot} illustrates a CAD model of the integration of the SwarmUS platform on one of the \pioneer. The \BBs are mounted on a custom 3D printed support\footnote{All the 3D models are open source and can be found in the main SwarmUS repository.} placing their UWB antennas 2.7 cm apart in a triangle formation.
Out of multiple conducted experiments to find the best antenna formation, this triangle pattern showed the best relative localization performances and is used throughout all this section. Fig. \ref{fig:robot} also shows the \HB supported by another 3D printed support beneath the Pioneer's lidar to save space and to lift the lidar over the \BBs antennas. 

Since the \turtlebot is a scalable layered robot, the integration of the SwarmUS platform on the \turtlebot required to add a new layer equipped with a \HB and \BB.  

A DC/DC converter was added on each robot to produce a 5 V bus from their 11.1 V LiPo battery. 
This converter powers directly the \HB, which then powers the three \BBs.
To reduce power consumption, the \HB automatically shutdowns the internal components of communication channels that aren't connected to a \BB.

In the \HB and three \BBs configuration, the SwarmUS platform consumes an average of 7 W while executing its Buzz script and localizing another platform. 
This measurement was taken by the displayed voltage and current value on an external power source powering the \HB at 5 V.

Each kit costs around 900 USD for, parts, cables and antennas included. The assembly cost is omitted since multiple assembly processes might be used to reduced cost or speed up the process.
However, the boards were designed in such a way that only the top side is populated by components, making assembly by hand easier. 

The following subsections describe tests that were performed to evaluate each feature provided by the SwarmUS platform.
\subsection{Localization}
\begin{figure}[!t]
    \centering
    \subfloat[Distance]{
        \includegraphics[width=0.23\textwidth]{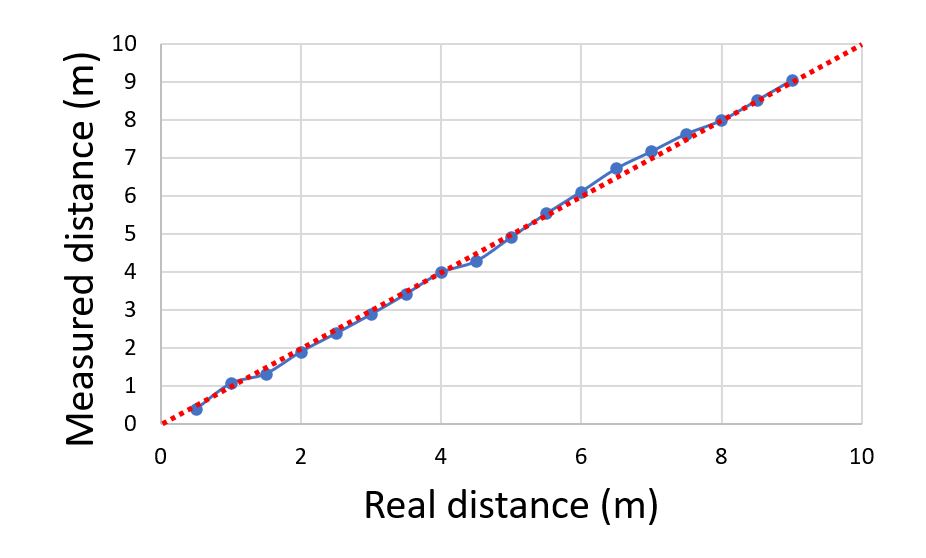}
        \label{fig:distance}
    }
    \subfloat[Angle]{
        \includegraphics[width=0.2\textwidth]{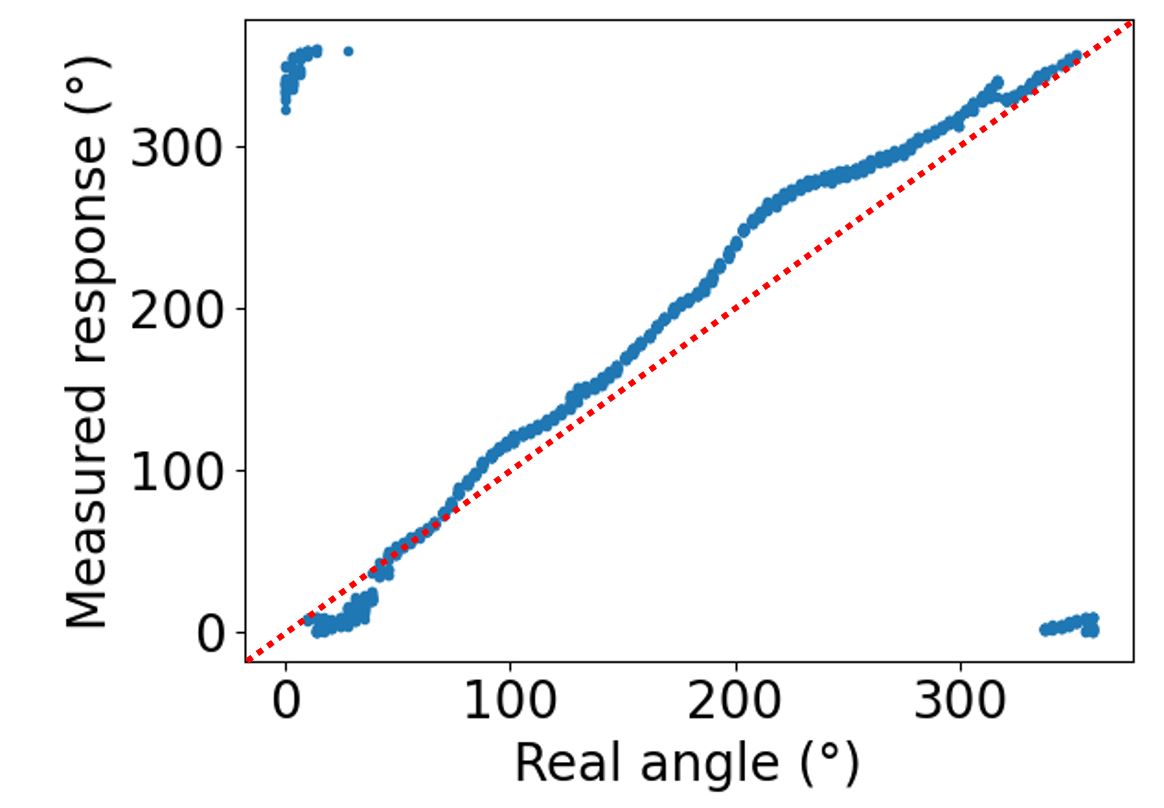}
        \label{fig:angle}
    }
    \hfill
    \subfloat[Standard deviation of the angle over 100 measures]{
        \includegraphics[width=0.2\textwidth]{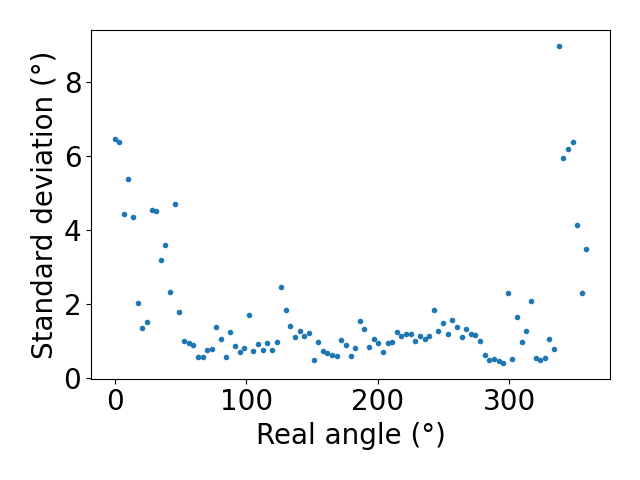}
        \label{fig:stdev}
    }
    \subfloat[Refresh rate of the devices]{
    \includegraphics[width=0.23\textwidth]{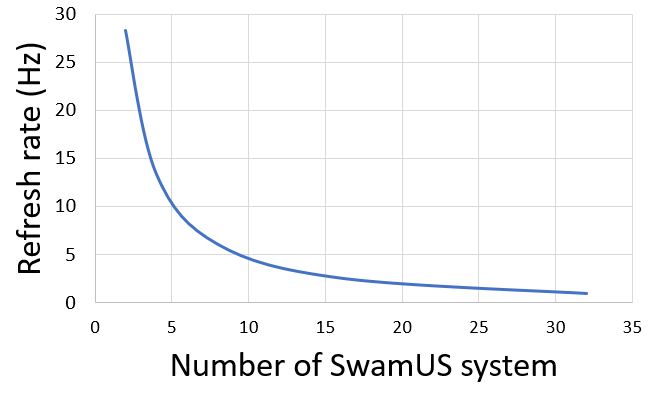}
    \label{fig:refresh}
    }
    \label{fig:localisation}
    \caption{Relative localization measurements}
\end{figure}

To characterize the relative localization system, multiple measurements of the position and bearing were made.
All the tests were made in a line-of-sight environment without any obstacle between the systems.
For the distance, 200 measures were taken every 50 cm from 0.5 m to 9 m (Fig. \ref{fig:distance}).
The average absolute error was 11 cm and the standard deviation was 3.43 cm.
For the bearing, a test bench composed of a turning table controlled by a step motor was created. An encoder at the output of the stepper motor output measured the bearing angle.
The table had step increments of 3.51$^{\circ}$ and was placed 2.5 m from a transmitting antenna.
For each step, 100 measures of the angle were made.
The average absolute error measured was 17$^{\circ}$ and the standard deviation was 1.7$^{\circ}$
(Fig. \ref{fig:angle} and \ref{fig:stdev}).

Since the localization system uses time frames to exchange information, the larger the swarm grows, the slower the refresh rate of the localization system will be, following a decreasing exponential rule.
The refresh rate depends on multiple factors such as the number of antenna used, the transmission speed, the speed of the SPI bus between the \MCU and DecaWave modules, the Rx and Tx timestamp size, the size of each frame individually and the idle time between each frame (Fig. \ref{fig:refresh}). 
It should be noted that the maximum number of robots 
configured for the localization system limits the refresh rate of measurements and not the actual number of robots. 


\subsection{Coordination}

A qualitative evaluation of the coordination system was performed with a "Follow the Leader" task.
This behavior was implemented by having a vector pushing the heading of the robot towards the leader. 
Although \textit{movebase} is used for obstacle avoidance, this ROS node cannot properly avoid other robots because they can't fully see each other with their lidar. 
For this reason, a second vector pushing robots from each other was added to avoid other robots.
The leader ID was maintained in the stigmergy to be changed dynamically.

At first, the validation was performed in a simulation of nine robots using the cross compiled version of the \HM in ROS and was used as a milestone before testing on real robots. The simulation was done with Gazebo, with ROS nodes and topics emulating the communication and relative localization systems. 
The simulation allowed to test the behavior of the swarm, independently of the hardware. 
The validation was executed by changing the leader of the swarm, at run time, using the simulated HSI from the Android Emulator. The simulated robots converged towards the dynamically changing leader as expected. However, some collisions occurred with nine robots, suggesting that the implemented Buzz behavior was not fully tuned.
Since the Buzz-based collision avoidance algorithm was relatively simplistic, a full integration with the navigation stack and proper obstacle avoidance could have given better results.


For real world validation, the same "Follow the leader task" was executed with two \pioneer and an handheld \HB and \BB assembly acting as the leader. 
One of the robot, without any other obstacles around it, was successfully able to follow the handheld leader, thus confirming that the platform was controlling the robot.


When testing with two \pioneer and the handheld leader, more emphasis was put on tuning the collision avoidance mechanism compared to the simulation. 
The imperfections of the relative localization system lead to different behaviors on the robots, mostly affected by the refresh rate. 
With a better mechanism for collision avoidance, where robots only avoid other robots in their heading direction,
a swarm of two robots were able to follow a third leader agent without collisions.
The Turtlebot was first included in the experimental swarm, but it wasn't successfully in achieving the task for mechanical integration reasons.

\subsection{Communication}

For the communication system, there are two important metrics: bandwidth and latency. The bandwidth limits the size and the frequency of updates on the swarm and the latency affects the frequency of updates due to the design of the communication system. Starting with the bandwidth, it limits the number of bytes per second that can be transmitted on the network. For a system running with Buzz, the most common used of the bandwidth is to share stigmergy updates between agents. The bandwidth usage is given by the following equation:

\begin{equation}
    Bandwidth = Size \cdot Frequency \cdot N^2 
\end{equation}

where Bandwidth is the a number of bytes transmitted per second in the whole network, Size is the amount of bytes per message used to share stigmergy updates, and $N$ the number of agents.
The bandwidth is proportional to the square of $N$ as each stigmergy in each robot is updated and propagated from that robot to the others.
Thus, the size of the stigmergy depends on the application as well as the required update frequency.
Measuring with iperf\footnote{https://iperf.fr/} in a  125 \xspace $m^2$ room, the lowest bandwidth measured was between 1.9 and 10.5 MB/s. This metric can be used as an indication of the feasibility of an application with a given stigmergy and swarm size. However, the bandwidth is heavily affected by distance, so measurements should be made in the target environment before deployment to validate the actual available bandwidth.

Concerning latency, due to the multilevel nature of the communication system, each message must pass by multiple fixed length queues before being sent.
For each level, for example between the two \MCUs or the network, there is a inherent latency between the insertion in the queue and the message being sent and then removed.
Pushing in this queue faster than the latency is a cause of message loss due to overflow.
The current highest latency being 9\xspace ms, the frequency of stigmergy updates is limited to approximately 100 Hz. This latency is introduced by the implementation of the communication protocol between the two \MCUs.

\subsection{Hardware Integration} \label{subsec:hardware_integration}
Since the SwarmUS platform aims to be installed on generic robots, it is important to validate how well it gets integrated on them.
First, since the ''Follow the Leader'' task has been executed and worked on the Pioneers, the whole system is functional: the \HM sends the desired messages to the \HMB which in return cause the desired actions of the robot. 
Also, the platform was adequately powered from the robot and didn't suffer from any power loss. 

However, mechanical issues appeared during the platform's integration on the \turtlebot. The weight added by the DC/DC converter, the \HB, three \BBs and their mechanical supports made the robot unable to cross obstacles as low as 1 mm without falling. 
On the other hand, the \pioneer, being bigger robots, had no stability or weight issues.

Another mechanical limit that has been observed with the space taken by the platform on the robot. 
The USB-C cables that go to the three \BBs can take around the double of the space taken by the board themselves. 
This unexpected occupied space is taken by the stiffer USB-C cables that are mechanically constrained to not follow a direct path between the \HB and its respective \BB. Placing other useful components for the robot in that space is difficult without adding worrying constraints on the USB-C cable.

Finally, the mechanical and electrical components close to the \BBs antennas changed the response of the localization system. 
For instance, on the TurtleBot, the Raspberry Pi computer and lidar were close to theses antennas and their interference in the localization system rendered a portion of the 0-360$^{\circ}$ circle unstable. 
On the Pioneer, the antennas were further away from other components and didn't suffer the same issue.


\section{DISCUSSION} \label{discussion}


The performance of the localization system can vary depending on the specific implementation.
As the standard deviation is lower than the average absolute error, the implementation of the each system could be significantly improved by correctly mapping the system response to experimental results. Furthermore, fusion of the relative localization measurements of the surrounding agents or even sensor fusion with the robots sensors could improve the precision of the system.
Changes in the environment and to the line-of-sight also changes the response of the system, increasing the absolute error average and the standard deviation. 
More tests should be conducted on the system to have a better understanding of how objects in the line-of-sight affects the system's response. 
However, it should be a good rule of thumb to install the \BBs somewhere on a robot so no components are near or potentially in its line line-of-sight.

The system also has a limitation regarding the scale of the swarm, like any other system that needs to share radio air space.
The refresh rate of the localization system exponentially decrease with each new system that wants to be localized, the maximum being 28 Hz for two robots.
Therefore, a smaller swarm can benefit from a fast refresh rate, where a bigger swarm might need to estimate theses distances and bearings by other means between each measurement.
A future improvement can be made to the scheduling mechanism by changing dynamically the number of time slots in the schedule to reflects the actual number of robots, thus always keeping the refresh rate to its optimal value.
It is also important to mention that the radio used for localization has a range of 9 m, other methods of localization could be used for longer distances.
On the other side, the actual limitation to the number of robots simultaneously using the localization system is also linked to the area they occupy, so the 9 m range become less limiting for larger robots.

Furthermore, some improvements could be made to the deployment mechanism of the Buzz script. 
Currently, when making changes to the Buzz script, the whole \HM firmware needs to recompiled and flashed to every \HB constituting the swarm.
This makes development and deployment on a large swarm a tiresome experience \cite{m_dorigo_swarm_2021}.
This could be alleviated with an Over-The-Air update system via the network to update the Buzz script of every agents in the swarm without having to individually flash them.

An important feature that showed promising results is the simulation of the whole platform in ROS. 
We only had access to three robots during the development of the platform, but the simulation capabilities enables the development of Buzz scripts on a bigger scale. 
The cross-compilation feature made the transition between the design and the real implementation effortless. 
To enhance the realism of the simulation, the imprecision of the relative localization could be added to the simulated measurements. 
Overall, the ROS integration and simulation of the platform is a key feature towards a standard platform.

Currently, the communication system relies on having one device managing the \wifi network for the entire swarm, which implies that the whole network can fail if this node is unreachable.
The communication system could be improved by having an election mechanism in the swarm to dynamically choose an agent as a router or integrate the ESP-WIFI-MESH library\footnote{https://www.espressif.com/en/products/sdks/esp-wifi-mesh/overview}, which would be relatively straightforward and improve the robustness of the platform. 
Additionally, the management of the messaging system could be optimized to reduce the 9 ms latency in each communication, thus increasing the bandwidth of the communication system.


On the hardware side, some areas of improvements could be made.
Firstly, a new integrated circuit by Decawave, the DW-3000, could upgrade the \BBs by simplifying the circuits around the UWB chip and rendering the \BBs more compact.
Secondly, the hardware used to transmit the data and the synchronized clock needed between the \BBs consumes around 35\% of the total power. 
This power consumption could be diminished by choosing better suited components for the application, a choice we did not have while the silicon crisis was in motion \cite{j_voas_scarcity_2021}.
Even if the total consumed power of 7 W is not an issue for mid-to-large size robots, the usage of lighter or even less components could diminish the power consumption without compromising the system accuracy as the clock distribution jitter is not the largest noise contributor in the distance and bearing measurements. 
For example, the redriver DS90LV001TLD/NOPB was added to overcompensate the signal integrity in the USB-C cable and could be removed without significant impact to the performance of the clock distribution system.
Lastly, components could be put on the underside of the boards, making the board much more compact. 
While it would complicate its assembly by hand, it would ensure a smaller footprint of the whole system.
For instance, the \turtlebot could benefit from such a reduction in the board footprint. 


These straightforward improvements to the original prototype would thus reduce cost, space and power consumption for the overall platform, making it attractive to larger scale production and adoption as a viable solution to the swarm hardware standardization problem.
The current design of the SwarmUS generic platform could fix three out of five key swarm robotics standards that needs to be established and followed by each swarm platform according to \cite{nedjah_review_2019}. 
Those standards are 1) a common minimum processing power, 2) a common communication interfaces and 3) a standard localization system that each swarm robotics platform should have. 
The SwarmUS platforms ensures those standards whatever the robotics platform because of its fixed hardware. 
However, further research should be made to ensure that the current performance of the SwarmUS system is sufficient. 
The two other standards according to \cite{nedjah_review_2019} are the standard sensors that a swarm platform should have and the standard tasks a swarm should be able to perform. 
We believe these standards to be dependent of the robot used and therefore out of scope for the SwarmUS platform alone.


\section{CONCLUSION}
Distributed swarm of robots shows promising features compared to centralized or single robots in the accomplishment of specific tasks. However, 
it has been demonstrated that the lack of standards around swarm robotics platform hinders their arrival as viable solution to real life problems \cite{nedjah_review_2019}.
This paper presents the SwarmUS platform has a potential solution to the standardization problem by including
many of 
the necessary features required by a swarm robotics platform on any model of robots. Even if the system has been successfully implemented on only two robots, the original hardware prototype performed correctly and showed promising results that could predict good performance of the platform on a larger swarm. Multiple solutions have been suggested to improve the prototype towards a generic swarm robotics platform
that can be adopted by the swarm robotics community at large.

\addtolength{\textheight}{-12cm}   




\section*{ACKNOWLEDGMENTS}
The SwarmUS project was supported by the
ASEQ foundation and the Université de Sherbrooke. 
Part of the work was also supported by a NSERC Discovery grant. 
The team wants to acknowledge the precious exchanges and advices given by Pr. François Michaud, Jonathan Bouchard and the \textit{Groupe de recherche en appareillage médical de Sherbrooke}  (GRAMS) laboratory. 




\bibliographystyle{IEEEtran}
\bibliography{SwarmUS}

\end{document}